\documentclass[lettersize,journal]{IEEEtran}
\usepackage{amsmath,amsfonts}
\usepackage{algorithmic}
\usepackage{algorithm}
\usepackage{array}
\usepackage[caption=false,font=normalsize,labelfont=sf,textfont=sf]{subfig}
\usepackage{textcomp}
\usepackage{stfloats}
\usepackage{url}
\usepackage{verbatim}
\usepackage{graphicx}
\usepackage{cite}
\usepackage{diagbox}

\usepackage[colorinlistoftodos]{todonotes}

\usepackage{soul}

\begin{document}
	
	\title{Using construction waste hauling trucks' GPS data to classify earthwork-related locations: A Chengdu case study}
	\author{Lei Yu$^1$ and Ke Han$^{2,*}$
    \thanks{This work is supported by the National Natural Science Foundation of China (no. 72071163) and the Natural Science Foundation of Sichuan Province, China (no. 2022NSFSC0474).}
    \thanks{$^{1}$L. Yu is with the School of Transportation and Logistics at Southwest Jiaotong University, Chengdu, China. $^{2}$K. Han is with the School of Economics and Management at Southwest Jiaotong University, Chengdu, China.}
\thanks{$^*$Corresponding author, email: kehan@swjtu.edu.cn}
    }
	\maketitle
	
	\begin{abstract}
		Earthwork-related locations (ERLs), such as construction sites, earth dumping ground, and concrete mixing stations, are major sources of urban dust pollution (particulate matters). The effective management of ERLs is crucial and requires timely and efficient tracking of these locations throughout the city. This work aims to identify and classify urban ERLs using GPS trajectory data of over 16,000 construction waste hauling trucks (CWHTs), as well as 58 urban features encompassing geographic, land cover, POI and transport dimensions. We compare several machine learning models and examine the impact of various spatial-temporal features on classification performance using real-world data in Chengdu, China. The results demonstrate that 77.8\% classification accuracy can be achieved with a limited number of features. This classification framework was implemented in the Alpha MAPS system in Chengdu, which has successfully identified 724 construction cites/earth dumping ground, 48 concrete mixing stations, and 80 truck parking locations in the city during December 2023, which has enabled local authority to effectively manage urban dust pollution at low personnel costs. 
	\end{abstract}
	
	
	\section{Introduction}\label{secIntro}

	\IEEEPARstart{U}{rban} dust pollution, primarily particulate matters (PM), is a major source of air pollution \cite{saad2019,souza2022,maues2021}. Its negative impact on air quality and public health is exacerbated by rapid urbanization and infrastructure development \cite{cheriyan2020review, li2010lca}. Earthwork-related locations (ERLs) are of particular interest to urban dust management as they are closely associated with the entire lifecycle of earthworks, from production (construction cites) to processing (concrete mixing stations), from transportation (truck parking spaces) to disposal (construction sites and earth dumping ground). Traditionally, these ERLs, in the number of several thousands in a city, are administered by various government branches depending on their categories, but their effective management is hindered by information delays, barriers, incorrectness, and incompleteness, which stem from manual inspection and bookkeeping. Aiming to enhance authorities' capability to manage urban ERLs in the digital age, this work proposes a machine learning framework to identify and classify ERLs using urban big data.
	
	Studies in earthwork management have mainly focuses on empirical assessment of environmental impact \cite{ghasemi2020experimental, hong2020empirical, yan2020systematic, luo2021occupational}, dust emission and dispersion patterns \cite{faber2015aerosol, gautam2015dispersion, ahmed2019emission}, and transport path planning \cite{li2022mixed, deng2020optimization}. These studies are concerned with specific ERLs with known locations and categories, but none have attempted city-wide identification and classification of massive amount of ERLs using digital technology and data intelligence, which this paper aims to address. The motivation is as follows: Maintaining a complete and up-to-date list of ERLs is crucial for urban \& environmental management, yet approximately 60\% of ERLs are not correctly registered with local authorities for reasons including inter-departmental information barriers, unauthorized earthworks, and insufficient manpower for on-site inspection. Moreover, different categories of ERLs are overseen by different government branches, and it is crucial to identify the ERLs' categories on a city scale to enable risk assessment and policy making on their parts. 
	
	In this work, we leverage big data analytics for efficient and intelligent sustainable urban management, by proposing a machine learning framework for ERLs classification. We introduce a comprehensive set of features, including basic geographic, land cover, points of interest, as well as transport related features extracted from the GPS data of over 16,000 construction waste hauling trucks (CWHTs). To our knowledge, this is the first attempt to classify ERLs for more effective urban management in the era of smart cities. Specific findings and contributions are as follows. 

	\begin{itemize}
	    \item  We employ four representative machine learning methods, namely, Logistic Regression (LR), Random Forest (RF), Gradient Boosted Decision Trees (GBDT), and Multilayer Perceptron (MLP). The results show that RF is the most effective, due to its superior ability to handle high-dimensional and unbalanced data, and its robustness to data outliers and noise.
	    \item We employ the SHAP method and feature distribution to extract importance of features. Additionally, we provide a detailed description of the relationship between features and different ERLs' categories, revealing the characteristics of each category while enhancing the interpretability of the machine learning results.
	    \item This study employs backward elimination to simplify the model features. The study's results demonstrate that a RF model, considering only a select few features, can achieve a high-level performance. This underscores the strong correlation between ERLs' categories and features such as location of ERLs, surrounding of ERLs, and stay time of CWHTs. 
	\end{itemize}

	The ERLs classification algorithm was successfully deployed in November 2023 in the {\it Alpha MAPS} system, for smart air pollution monitoring and management in Chengdu. During December 2023, the module has performed classification tasks for 16,132 ERLs. A random sampling based on on-site verification suggests a classification accuracy of 77.8\%. The ERLs classification functionality has considerably enhanced local authorities' capacity to monitor and manage locations with high dust pollution risks. 

	This paper is organized as follows: Section \ref{sec:Literature} provides a comprehensive review of relevant research progress; Section \ref{sec:Data} introduces concepts related to the problem and a nearly comprehensive set of features that could impact the categories of ERLs; Section \ref{sec:model} describes the classification model, performance metrics, and feature importance identification employed in this study; subsequently, Sections \ref{sec:results model} and \ref{sec:results feture} present the computational results and the corresponding discussion; and finally, Section \ref{sec:system} presents the system deployment and Section \ref{sec:conclusion} presents the conclusions. 

	\section{Literature review}\label{sec:Literature}

	Large-scale urban development is often closely related to the construction industry and is one of the main sources of dust pollution \cite{cheriyan2020review}. According to our survey, research on earthwork pollution has raised extensive concerns regarding urban dust pollution management and intelligent transportation systems. Research on the management of urban dust pollution has primarily focused on dust pollution evaluation models \cite{ghasemi2020experimental, hong2020empirical, yan2020systematic, luo2021occupational} and dust emission patterns \cite{faber2015aerosol, gautam2015dispersion, ahmed2019emission}. Research on intelligent transportation systems has predominantly focused on traffic path planning \cite{li2022mixed, deng2020optimization} for CWHTs. Although most of these studies examine dust pollution during transportation, the transportation process entails numerous uncertainties, and assessing its risks relies on understanding the sources of earth-moving pollution. Consequently, the identification and classification of ERLs have become increasingly crucial for practical applications. Traditional methods rely primarily on manual processes, such as random sampling, which are labor-intensive and inefficient. Employing big data and artificial intelligence to facilitate the management of earthwork pollution is crucial.

	The investigation of the sources of urban dust pollution also encompasses the field of big data mining. \cite{mayer2013big} describes big data technology as ``activities that can be performed on a large scale, impossible at smaller scales, to generate a novel form of value." Numerous researchers have embraced three definitions of Big Data, namely volume, variety, and velocity, often referred to as the ‘three Vs' \cite{mcafee2012big}. Big Data analytics can reveal concealed patterns, unknown correlations, and other valuable insights \cite{lu2019big}. \cite{2023HanChenCaiyun} have identified stay points from GPS trajectory points of CWHTs using an adaptive spatio-temporal threshold method, while \cite{2024HanWangyebing} have extracted long-term and regular earthwork point information using a grid clustering algorithm based on this data. By contrast, \cite{bi2022identification} have identified hot nodes in CWHTs operations using Xgboost and DBSCAN. \cite{lu2019big} have classified a specific operational pattern, namely illegal dumping. However, few studies have classified nodes related to earthwork. Nevertheless, note that \cite{ji2022precision} have identified and classified areas of hazardous material operations using trajectory data from hazardous material trucks. Our study aims to classify ERLs, thereby addressing a significant research gap in this field.

	Big data and machine learning algorithms are closely intertwined. Machine learning algorithms are commonly classified into four categories: supervised, unsupervised, semi-supervised, and reinforcement learnings \cite{mohammed2016machine}. Supervised learning involves learning a function that maps inputs to outputs using input-output pairs of samples \cite{han2022data}. Supervised learning tasks typically encompass classification, which addresses discrete labeled data, and regression, which handles continuous labeled data. The data used for supervised learning may be structured or unstructured. Our research approach concentrates on multi-classification models utilizing structured data, employing commonly used algorithms, such as Naive Bayes, Linear Discriminant Analysis, Logistic Regression (LR), K-nearest neighbors (KNN), Support Vector Machine (SVM), Random Forest (RF), Adaptive Boosting (AdaBoost), and Multilayer Perceptron (MLP) \cite{han2022data, witten2002data}. Model performance and interpretability are essential. However, machine learning models often lack interpretability, leading to the increased adoption of the SHAP method in research \cite{lundberg2017unified, lundberg2018consistent}. The Shapley value, which is based on cooperative game theory, was originally proposed by Shapley in 1953 \cite{shapley1953value}. In 2017, \cite{lundberg2017unified} developed a Python package capable of computing SHAP values for various models, including Multilayer Perceptron (MLP) and Random Forest (RF), thereby simplifying the interpretation of machine learning models.

	The literature review reveals significant interest among researchers regarding urban dust pollution. However, there have been no systematic investigations on the classification of ERLs. In this study, we utilize CWHTs' GPS trajectory , POI, and land cover data to employ various multi-classification machine learning methods to achieve highly accurate prediction results. In addition, we utilize the advanced SHAP method to elucidate the machine learning model. This study addresses the research gap in the classification of ERLs.

	\section{Data}\label{sec:Data}

\begin{figure}[!t]
	\centering
	\includegraphics[width=0.45\textwidth]{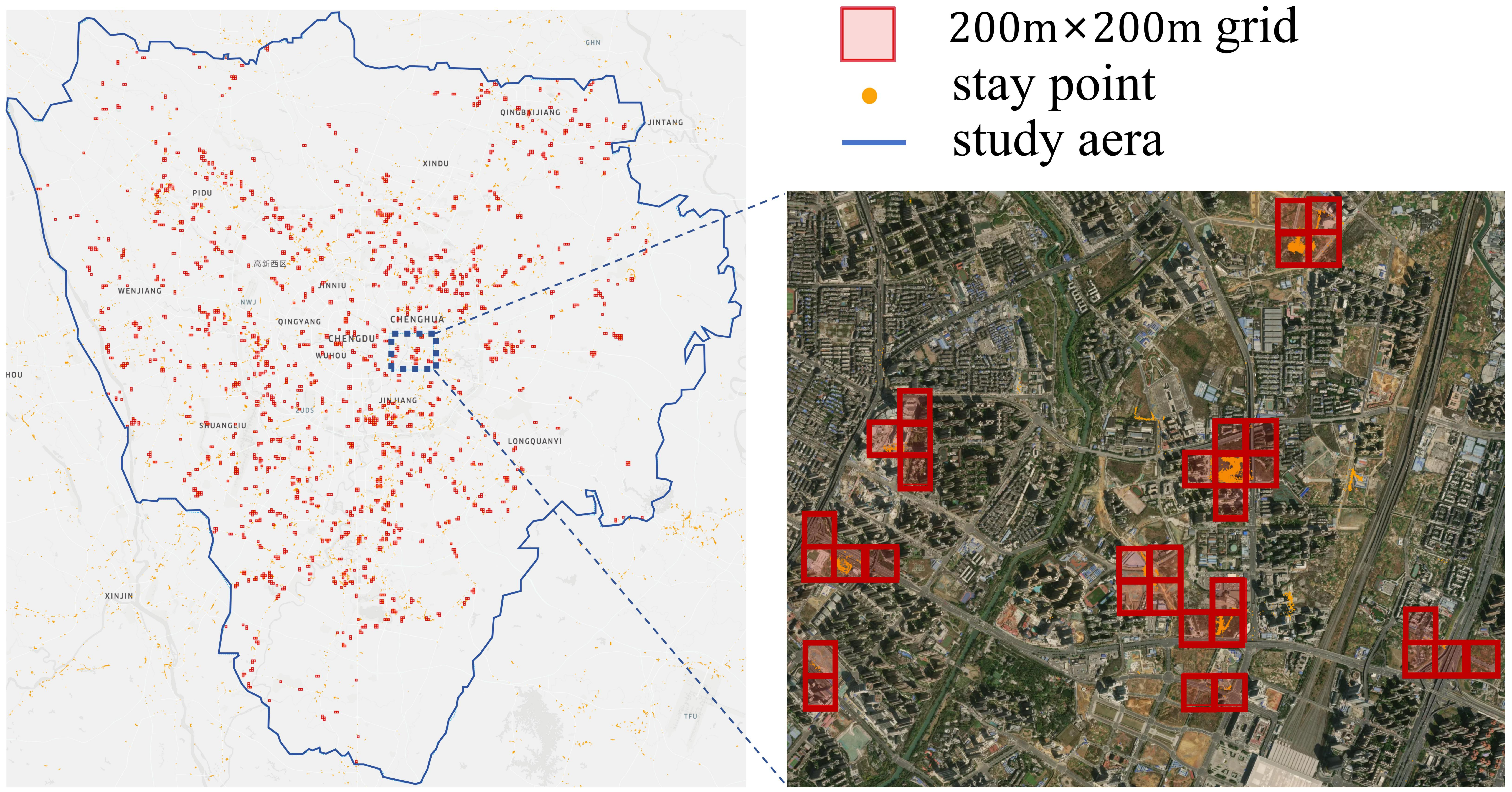}
	\caption{\label{fig:ERLs_geo}Spatial distribution of ERLs, (Day shift June 5, 2023). Each ERL is represented as a group of small grids of size 200m$\times$200m}
	\end{figure}

	Chengdu is a major city with a population of over 20 million. Its core area, spanning over 3,700 km$^2$, is chosen for the study. The ERLs are identified as follows. (1) From the full GPS trajectory points, we identify a subset that are associated with full stop or low speeds, which are referred to as {\it stay points}. This is achieved via a threshold of displacement within certain time interval, as detailed in \cite{yang2023}. (2) A multi-layer clustering approach proposed by \cite{2023HanChenCaiyun} and \cite{2024HanWangyebing} is employed to identify stay point clusters that appear regularly in the study area, indicating earthwork-related locations. (3) These clusters are mapped to small grids of size 200m$\times$200m, forming pixelated representation of ERLs with irregular shapes and varying sizes throughout the city. As illustrated in Fig. \ref{fig:ERLs_geo}, each ERL is represented as a group of small grids. 
	
	We classify ERLs into three categories: Earthwork\_Related (ER), Material\_Related (MR), and Parking \& Maintenance (PM) (Table~\ref{tab:category}). ER includes construction sites and earth dumping ground, MR includes quarry and concrete mixing stations, PM includes parking, auto repair and maintenance stations.

\begin{figure}[!t]
	\centering
	\includegraphics[width=0.45\textwidth]{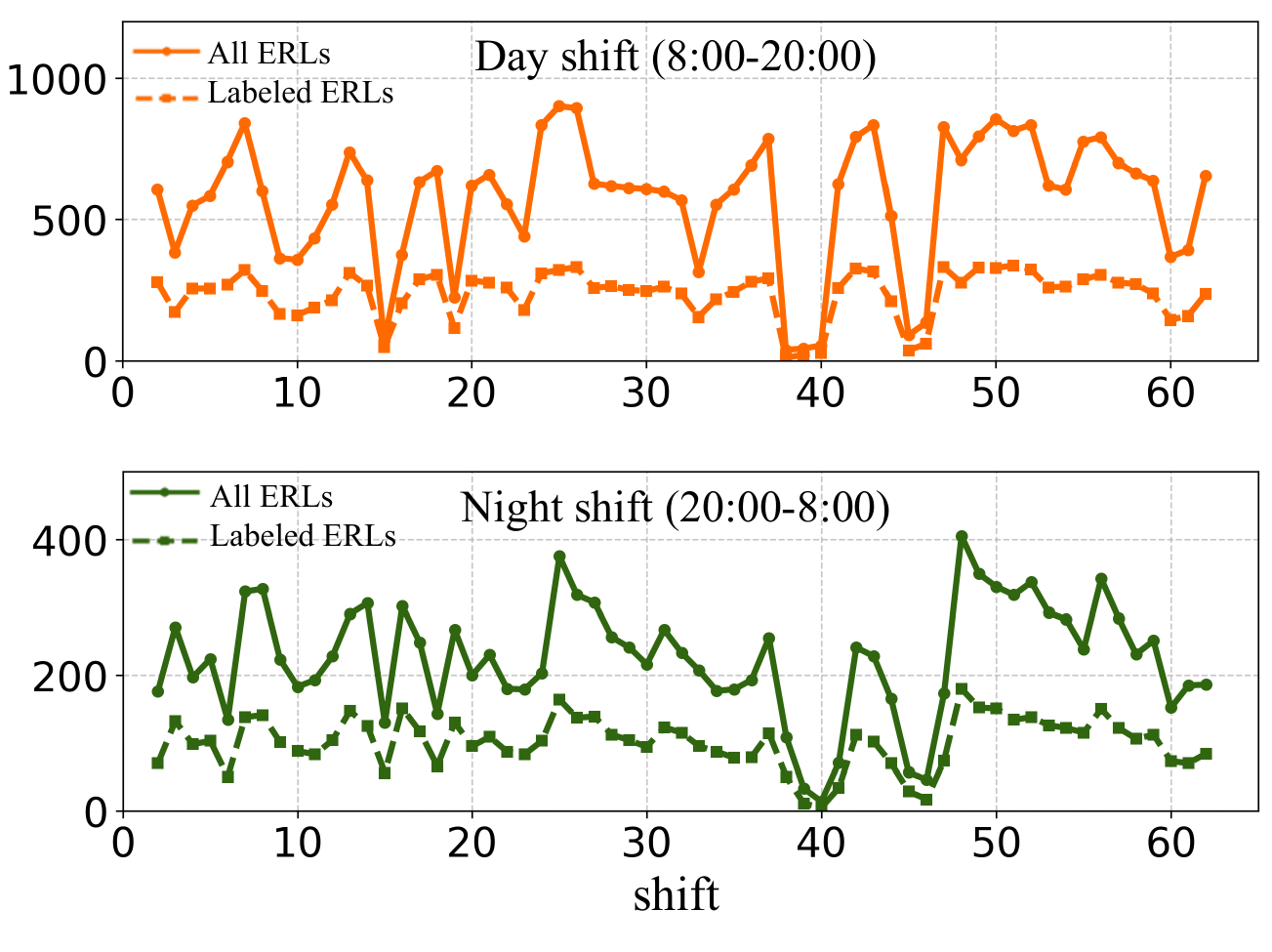}
	\caption{Number of ERLs with CWHTs operation during each shift. (Every 12-hr period is defined as one shift and the dates are from 1 May to 31 June 2023.)}
	\label{fig:ERLs_num}
	\end{figure}

	We take the study period from 1 May to 31 June 2023, with a total of 61 days. Every 12 hours (8:00 to 20:00, or 20:00 to 8:00 the next day) is defined as a shift, totaling 122 shifts. An ERL is called a {\it sample} during a shift if there were CWHT activities. During the 122 shifts there were 52,536 ERL samples, whose temporal distributions are shown in Fig. \ref{fig:ERLs_num}. Of these ERL samples, a total of 20,489 (39\%) have known categories or `labels' (including 13,401 ERs, 1,559 MRs and 5,529 PMs), obtained either from official registration or on-site inspection. On the one hand, these labeled samples are sufficient for machine learning; on the other hand, about 61\% of ERL samples have unknown categories, necessitating the classification of ERLs for effective urban environmental management.

	It is important to acknowledge the presence of sample imbalance in this dataset, which could potentially impact model performance \cite{johnson2019}. This issue may be addressed by three different approaches: data-level techniques, algorithm-level methods, and hybrid approaches \cite{krawczyk2016}. However, developing a specialized method to address this imbalance is deemed unnecessary for two main reasons: firstly, the majority of samples belong to the ER category, which are the main targets of environmental management. Secondly, when it comes to minority samples, their number outweigh the relative proportion \cite{krawczyk2016}. In our case, the smallest category has 1,559 samples, which is sufficient. Therefore, we adopt an algorithm-level approach, namely the cost-sensitive learning method, which achieves good results (see Section \ref{sec:results model}). This method is easy to implement, adds no extra computational burden, and has been proven effective in many instances \cite{johnson2022, mienye2021, hoppner2022}.

\begin{figure*}[!t]
	\centering
	\includegraphics[width=0.95\textwidth]{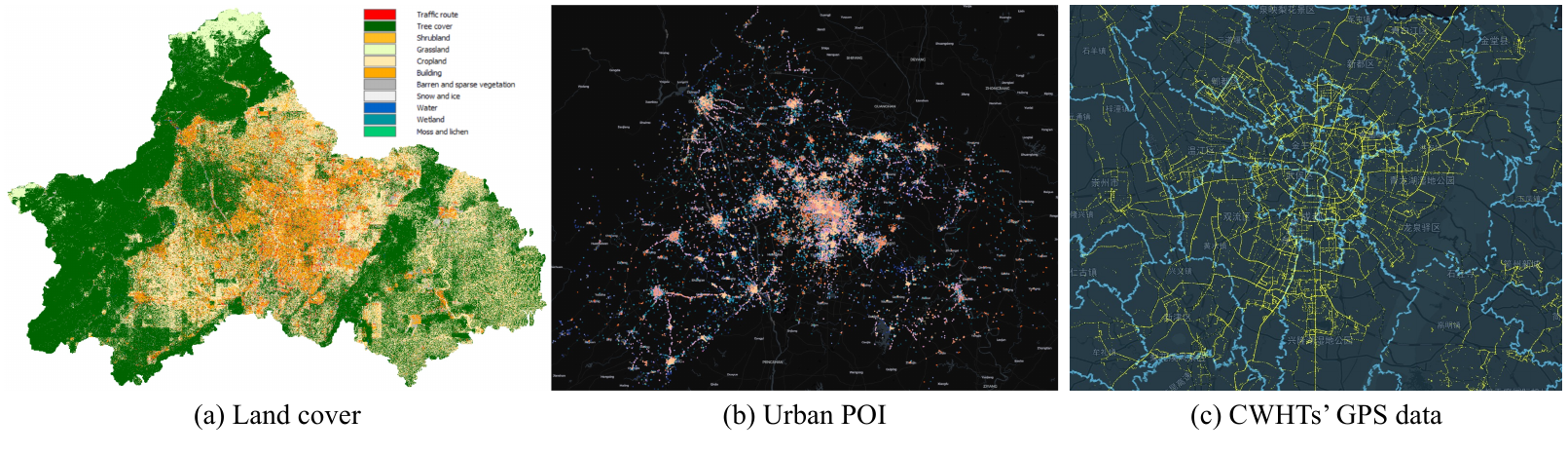}
	\caption{\label{fig:Data} Raw data used to generate various features.}
	\end{figure*}

	To accurately classify the ERLs, we thoroughly considered all features that may be relevant to the ERLs' functionality or characteristic. In particular, 58 numerical features are employed, encompassing four primary aspects: basic geographic, land cover (Fig. \ref{fig:Data} (a)), urban POI (Fig. \ref{fig:Data} (b)), and transport related features. Where basic geographic and transport related features are extracted from the GPS data of over 16,000 CWHTs (Fig. \ref{fig:Data} (c)).
	
	\begin{table}[!t]
	    \centering
	    \caption{The location category and the corresponding risk level.}
	    \label{tab:category}
	    \begin{tabular}{l l l l}
	        \hline
	        Category & Stay locations \\
	        \hline
	        ER & construction sites, earth dumping ground \\
	        MR & Quarry, concrete mixing station & \\
	        PM & Parking, auto repair \& maintenance stations \\
	        \hline
	    \end{tabular}
	\end{table}
	
	\subsection{Basic geographic features}
	The basic geographic features are related to the structure and location of each ERL. Table~\ref{tab:geographic} lists the four features. Specifically, $distance\_LR$  denotes the distance between the left and right boundaries. $distance\_UL$  represents the upper and lower boundaries of the sample region. $num\_grid$ indicates the number of $200m\times200m$ grids in each sample region. $distance\_LR$, $distance\_UL$, and $num\_grid$ encode information on the approximate shapes and sizes of the ERLs. Additionally, $distance\_center$ represents the Euclidean distance between the geometric center of the ERL and the projected coordinates of the center of Chengdu (Tianfu Square). The feature $distance\_center$ is partly related to the level of urbanization. 

	\begin{table}[!t]
	    \centering
	    \caption{Geographic features.}
	    \label{tab:geographic}
	    \begin{tabular}{l l}
	        \hline
	        Features & Description \\
	        \hline
	        $num\_grid$ & Number of grids \\
	        $distance\_LR$ & Distance between left and right boundaries \\
	        $distance\_UL$ & Distance between upper and lower boundaries \\
	        $distance\_center$ & Distance from city center \\
	        \hline
	    \end{tabular}
	\end{table}
	
	\subsection{Land cover features}
	
	Satellite remote-sensing data have achieved good results in the field of location identification and classification \cite{chen2020classification, lynch2020classification}. To classify ERLs and reduce data processing costs, we incorporate structured land cover data into the feature set. Table~\ref{tab:land cover} lists eight common land cover features obtained from \cite{li2023sinolc}, with a data resolution of 1 m. To mitigate the influence of the ERL size on the classification results, we calculate the area proportions of the eight land cover features as the final feature. The areas of different ERLs within the same category may vary significantly, potentially resulting in differences of orders of magnitude in land cover features.
	
	\begin{table}[!t]
	    \centering
	    \caption{Land cover features.}
	    \label{tab:land cover}
	    \begin{tabular}{l l}
	        \hline
	        Features & Description \\
	        \hline
	        $traffic\_route$ & Area ratio of transportation facilities\\
	        $tree\_cover$ & Area ratio of trees\\
	        $grassland$ & Area ratio of low herbaceous\\
	        $cropland$ & Area ratio of arable land and human planted crops\\
	        $building$ & Area ratio of human-made building\\
	        $barren\_S\_V$ & Area ratio of barren and sparse vegetation\\
	        $water$ & Area ratio of waters\\
	        $moss\_lichen$ & Area ratio of moss and lichen\\
	        \hline
	    \end{tabular}
	\end{table}
	
	\subsection{POI features}

	Urban POI data are frequently used as features for urban location identification and classification \cite{zheng2023identification, luo2023urban}. POI can be categorized in various ways; we divide them into 17 categories, as listed in Table~\ref{tab:POI}. We calculate the total number of each category of POI within a one-kilometer radius of the geometric center of each ERL, where $all\_poi$ represents the sum of all the POI.

    \begin{table}[!t]
        \centering
        \caption{Features of POI.}
        \label{tab:POI}
        \begin{tabular}{l l}
            \hline
            Features & Description \\
            \hline
            $food$ & Number of food services within 1km radius\\
            $road\_facility$ & Number of road facilities within 1km radius\\
            $scenic\_spots$ & Number of scenic spots within 1km radius\\
            $public$ & Number of public facilities within 1km radius\\
            $enterprise$ & Number of enterprises within 1km radius\\
            $shopping$ & Number of shopping services within 1km radius\\
            $trans\_facility$ & Number of transportation facilities within 1km radius\\
            $financial$ & Number of financial services within 1km radius\\
            $science\_E$ & Number of education services within 1km radius\\
            $trans\_F$ & Number of transportation facilities within 1km radius\\
            $car\_S$ & Number of car services within 1km radius\\
            $car\_R$ & Number of car repair services within 1km radius\\
            $business$ & Number of business residences within 1km radius\\
            $subsistence$ & Number of subsistence services within 1km radius\\
            $sports$ & Number of sporting services within 1km radius\\
            $health$ & Number of health care services within 1km radius\\
            $government$ & Number of government facilities within 1km radius\\
            $accom$ & Number of accommodation services within 1km radius\\
            $all\_poi$ & Number of all POI within 1km radius\\
            \hline
        \end{tabular}
    \end{table}

	\subsection{Transport related features}\label{sec:Transport related features}
	ERLs are identified based on the trajectories of CWHTs; therefore, transport related features such as truck flow and stay points may affect classification accuracy. For instance, PM exhibits a higher outflow during the morning peak and a higher inflow during the evening peak, unlike ER and MR. ER and MR prioritize operational efficiency with shorter stay times (dwell times) for CWHTs, whereas PM experiences longer stay times.
	Based on CWHTs trajectory data, we calculate 20 transport related features presented in Table~\ref{tab:Transport}. Here, $flow\_t$ represents the inflow of the ERL during each hour $t$ and $all\_flow$ represents the total flow over the span of 12 h. Meanwhile, $stay\_t$ represents the cumulative number of stay points within the ERL area for each hour $t$ $(t=1,2,\dots,12)$, and $all\_stay$ represents the total number of stay points over the course of 12 h. Although the physical meaning of $stay\_t$ is not as explicit as $flow\_10$, it may contain additional information. $degree$ represents the sum of the in- and out-degrees of each ERL, where the degree is the number of edges directly connected to a node. $stay\_time$ represents the average stay time for all the CWHTs remaining at a specific ERL.

\begin{table}[!t]
    \centering
    \caption{Features of Transport related features.}
    \label{tab:Transport}
    \begin{tabular}{l l}
        \hline
        Features & Description \\
        \hline
        $stay\_t$ & Number of stay points per hour ($t=1,2,\dots,12$)\\
        $all\_stay$ & Total number of stay points\\
        $flow\_t$ & Number of in-flow per hour ($t=1,2,\dots,12$)\\
        $all\_flow$ & Total number of in-flow\\
        $degree$ & Number of in-degrees and out-degrees\\
        $stay\_time$ & Average length of stay\\
        \hline
    \end{tabular}
\end{table}

	\section{Classification models and feature importance}\label{sec:model}
	
	In this section, four classification models are introduced for classifying ERLs, and five performance metrics are adopted to address and compare the model performances. Additionally, the SHAP method is introduced for model interpretation, aiming to identify feature importance and demonstrate high classification performance on a more limited subset of features.
	
	\subsection{Model}
	
	The selection of prediction models can influence prediction performance \cite{wang2021aircraft}. We introduce four machine-learning models for the ERL categories: MLP, LR, GBDT, and RF.
	
	\subsubsection{Multilayer perceptron}
	
	The MLP \cite{han2022data} is a   of feedforward neural network with multiple layers. It comprises an input layer, one or more hidden layers, and an output layer. Each layer is composed of multiple neurons that are connected to all the neurons in the preceding layer. Typically, neurons other than the input neurons use nonlinear activation functions, aiding the network in learning more complex mapping relationships. Among them, the commonly used Relu activation function is constructed as
	\begin{equation}
	\text{Relu}(v) = \max(0, v)
	\end{equation}
	where $\text{Relu}(v)$ denotes the output and $v$ denotes the input.
	
	For multi-classification problems, the Softmax function is commonly employed for the output layer. The function maps the values of the output neurons to a probability distribution, ensuring that the probability value for each category falls between zero and one and that the sum of the probabilities for all categories is one. The mathematical expression for the Softmax function is 
	
	\begin{equation}
	\text{Softmax}(\mathbf{z}_i)=\frac{e^{z_i}}{\sum_{i}^{K} e^{z_j}}\label{equ:Softmax}
	\end{equation}
	where $\mathbf{z}$ is the input vector, and $K$ is the number of categories.
	
	The cross-entropy loss function \eqref{equ:cross-entropy} is commonly used for multi-classification problems during training. The model parameters are trained using a widely recognized backpropagation algorithm \cite{han2022data}.
	
	\begin{equation}
	J(\theta)=-\frac{1}{N}\sum_{i=1}^{N}\sum_{j=1}^{K} y_{ij}log(\hat{y}_{ij})\label{equ:cross-entropy}
	\end{equation}
	where $N$ represents the number of samples, $K$ is the number of categories, $y_{ij}$ is the true label (one-hot coding) of sample $i$, and $\hat{y}_{ij}$ is the model's predicted probability that sample $i$ belongs to category $j$.
	
	\subsubsection{Logistic regression}
	
	LR is a generalized linear model commonly used to address binary classification problems \cite{han2022data}. Softmax functions can be introduced into the output layer to further expand the generalized LR to Softmax regression, which is suitable for resolving multi-classification problems. Eq~\eqref{equ:linear} represents a generalized LR. The model employs a Softmax function (Eq~\eqref{equ:Softmax}) at the output layer to translate the raw scores of each category into probabilities, and then determines the model loss based on the cross-entropy loss function (Eq~\eqref{equ:cross-entropy}):
	
	\begin{equation}
	z_j=\mathbf{w}_j\mathbf{x}+b_j\label{equ:linear}
	\end{equation}
	where $\mathbf{x}$ represents the input feature vector, and $\mathbf{w}_j$ and $b_j$ are the model coefficients. $z_j$ is the output of the linear regression predicting the $j$ category.
	
	\subsubsection{Gradient Boosting Decision Trees}
	
	GBDT is a versatile nonparametric statistical learning approach extensively used for multi-class classification tasks. The modeling process unfolds in a stepwise manner, enabling the optimization of any differentiable cost function \cite{ke2017lightgbm}. 
	Similar to other boosting techniques, gradient boosting integrates weak learners iteratively to create a robust learner. By introducing estimator $h$, the refined prediction model is derived as $F_{s+1}(x)=F_s(x)+h(x)$. The estimator $h$ is selected as $h(x)=y-F_s(x)$, where $y$ is the actual output value.
	
	The GBDT has gained widespread popularity owing to its efficiency, accuracy, and interpretability \cite{ke2017lightgbm}. It is a promising method for precisely tackling multi-class classification tasks. The GBDT constructs a series of decision trees and statistical models that are tailored for supervised prediction challenges. These trees make predictions through a sequence of decisions outlined in the tree structure, with each node representing a split in the potential values for a specific feature. The utility of decision-tree regression lies in its ease of interpretation and visualization. Moreover, it has the potential to reveal patterns that may be challenging to identify using traditional regression methods. This makes the GBDT a robust choice for efficiently and accurately tackling multi-class classification challenges.
	
	\subsubsection{Random forest}
	
	The RF is an ensemble prediction model composed of numerous decision trees. Remarkably, RF often yields favorable results, even without extensive hyperparameter tuning \cite{bernard2009influence}. During the training process, each tree in the RF learns from a randomly sampled subset of the data using a bagging technique. Notably, the samples are repeatedly applied within a single tree, resulting in the entire forest having a lower variance without a corresponding increase in bias. RF predictions are derived by averaging the predictions of each decision tree within the ensemble. 
	
	Although individual decision trees explore the splits for each feature in every node, RF focuses on a split for only one feature per node. Initially, a small subgroup of explanatory features is randomly selected. Subsequently, the node is split using the best feature from the limited set, and a new set of eligible features is arbitrarily selected. This iterative process continues until the tree is fully grown. Ideally, each terminal node contains only one observation. As the number of features increases, the eligible feature set may differ significantly from one node to another. However, prominent features eventually emerge in the tree, and their respective predictive success enhances the overall reliability.
	
	\subsection{Performance metrics}
	
	In this section, we compute the confusion matrix (Table~\ref{tab:Confusion}), where True Positive (TP) denotes the model correctly classifying positive examples as positive, False Positive (FP) represents the model incorrectly classifying negative examples as positive, True Negative (TN) indicates the model correctly categorizing negative cases as negative, and False Negative (FN) signifies the model incorrectly classifying positive examples as negative. Subsequently, we present five predictive performance metrics to assess and compare the performance of the predictive models.
	
	\begin{table}[!t]
	    \centering
	    \caption{Confusion Matrix.}
	    \label{tab:Confusion}
	    \begin{tabular}{|c|c|c|}
	        \hline
	        \diagbox{Reference}{Prediction} & Positive & Negative \\
	        \hline
	        Positive & TP & FN \\
	        \hline
	        Negative & FP & TN \\
	        \hline
	    \end{tabular}
	\end{table}
	
	\subsubsection{Accuracy}
	
	Accuracy quantifies the overall proportion of correct predictions made by the model, as shown in Eq~\eqref{equ:Accuracy}. In situations of category imbalance, the accuracy may be inappropriate, as the model may exhibit a bias towards predicting a higher number of categories.
	\begin{equation}
	\text{Accuracy}=\frac{TP+TN}{TP+TN+FP+FN}\label{equ:Accuracy}
	\end{equation}
	
	\subsubsection{Precision}
	
	Precision quantifies the proportion of samples predicted to be positive instances that are positive instances by the model, as depicted in Eq~\eqref{equ:Precision}. Precision is highly effective for addressing issues that aim to minimize FP instances, such as spam categorization. However, it exhibits less robustness in noisy situations because it concentrates solely on accuracy in positive examples.
	\begin{equation}
	\text{Precision}=\frac{TP}{TP+FP}\label{equ:Precision}
	\end{equation}
	
	\subsubsection{Recall}
	
	Recall quantifies the proportion of samples that the model successfully predicts as positive among the positive samples, as illustrated in Eq~\eqref{equ:recall}. Recall is valuable when addressing the challenge of minimizing FN cases, such as in cancer diagnoses. However, in noisy situations, the occurrence of FPs may increase.
	\begin{equation}
	\text{Recall}=\frac{TP}{TP+FN}\label{equ:recall}
	\end{equation}
	
	\subsubsection{F1-score}
	
	The F1-score combines precision and recall and serves as a reconciled average of the two. In the dichotomous case, the F1-score is typically calculated for the category of interest (usually the positive category), and its formula is shown in Eq~\eqref{equ:F1}. In the multi-category case, three variants exist: Macro-F1, Micro-F1, and Weighted-F1. Because of the imbalance in the data samples in this study, it is suitable to utilize Macro-F1 for a comprehensive evaluation of the model's performance, as outlined in Eq~\eqref{equ:Macro-F1}.
	\begin{equation}
	\text{F1-score}=\frac{2 \times Precision \times Recall}{Precision+Recall}\label{equ:F1}
	\end{equation}
	
	\begin{equation}
	\text{Macro-F1}=\frac{1}{K}\sum_{i=1}^{K}F1\text{-}scrore_i\label{equ:Macro-F1}
	\end{equation}
	where $K$ is the total number of categories and $F1\text{-}score_i$ is the F1-score of the $kth$ category.

	\begin{table*}[!t]
		\centering
		\caption{Comparison of the performance of five evaluation metrics for ERL classification is presented, including mean and standard deviation. The best performance is highlighted in bold.}
		\label{tab:Performance}
		\begin{tabular}{c c c c c c}\hline
			Model & Accuracy & Precision & Recall & Macro-F1 & AUROC\\ \hline
			LR & 0.663$\pm$0.007 & 0.483$\pm$0.016 & 0.427$\pm$0.008 & 0.429$\pm$0.011 & 0.722$\pm$0.009\\
			MLP & 0.756$\pm$0.017 & 0.754$\pm$0.016 & 0.568$\pm$0.019 & 0.597$\pm$0.023 & 0.837$\pm$0.012\\
			GBDT & 0.778$\pm$0.012 & \textbf{0.814$\pm$0.019} & 0.541$\pm$0.015 & 0.583$\pm$0.026 & 0.862$\pm$0.013\\
			RF & \textbf{0.785$\pm$0.014} & 0.803$\pm$0.018 & \textbf{0.569$\pm$0.017} & \textbf{0.609$\pm$0.021} & \textbf{0.870$\pm$0.011}\\
			\hline
		\end{tabular}
	\end{table*}

	\subsubsection{AUROC}
	
	The receiver operating characteristic (ROC) curve evaluates the performance of the classification model across various classification thresholds. The curve represents the True Positive Rate (TPR, as shown in Eq~\eqref{equ:TPR}) on the Y-axis and the False Positive Rate (FPR, as shown in Eq~\eqref{equ:FPR}) on the X-axis. Proximity to the upper-left corner indicates better model performance, with no impact of sample imbalance on the ROC curve. The area under the curve (AUC) represents the area beneath the ROC curve; the closer the AUC is to 1, the better the model performance.
	\begin{equation}
	TPR=\frac{TP}{TP+FN}\label{equ:TPR}
	\end{equation}
	\begin{equation}
	FPR=\frac{FP}{FP+TN}\label{equ:FPR}
	\end{equation}
	
	\subsection{Feature importance}\label{sec:Feature importance}
	
	The SHAP method, which is rooted in cooperative game theory and based on Shapley values, serves as an approach for interpreting machine learning models \cite{lundberg2017unified}. SHAP represents the output model as a linear summation of the input variables. Assuming that the input variable of the model is $\mathbf{x}$, the interpretive model $g(\mathbf{x}^{'})$ of the original model $f(\mathbf{x})$ is denoted by
	
	\begin{equation}
	f(\mathbf{x})=g(\mathbf{x}^{'})=\phi_0+\sum_{i=1}^{M}\phi_ix_i^{'}\label{equ:SHAP1}
	\end{equation}
	where $\mathbf{x}^{'}$ represents the simplified input, $M$ represents the number of input features, and $\phi_0$ represents the output value of the model when all the inputs are missing. The inputs $\mathbf{x}^{'}$ and $\mathbf{x}$ are related by the mapping function $\mathbf{x} = h_x(\mathbf{x}^{'})$. Eq~\eqref{equ:SHAP1} has a unique solution and exhibits three ideal properties: local accuracy, missingness, and consistency. The only possible model that satisfies these properties is:
	
	\begin{equation}
	\phi_i(f,\mathbf{x})=\sum_{\mathbf{z}^{'}\subseteq \mathbf{x}^{'}} \frac{| \mathbf{z}^{'} |!(M-| \mathbf{z}^{'} |-1)!}{M!}[f_x(\mathbf{z}^{'})-f_x(\mathbf{z}^{'}\backslash i)]\label{equ:SHAP2}
	\end{equation}
	where $| \mathbf{z}^{'} |$ represents the number of non-zero entries in $\mathbf{z}^{'}$. $\phi_i$ is the Shaply value from Eq~\eqref{equ:SHAP1} and $\mathbf{z}^{'}\backslash i$ represents $z_i^{'}=0$. $f_x(\mathbf{z}^{'})=f(h_x(\mathbf{z}^{'}))=E[f(\mathbf{z})|\mathbf{z}_s]$ and $S$ is the set of nonzero indices in $\mathbf{z}^{'}$, known as SHAP values.
	
	SHAP provides effective explanations for both local and global model interpretations. In this study, we employ the method proposed by \cite{lundberg2018consistent} for Shapley value calculations tailored to tree models, reducing the computational complexity from $O(TL2^M)$ to $O(TLD^2)$ where $T$ is the number of trees, $L$ is the maximum number of leaves in any tree, $M$ is the number of features, and $D$ is the maximum depth of any tree. Calculating feature importance serves two purposes: to analyze the contribution of each feature to predictions for model interpretation, feature selection, and performance improvement; and to analyze ERL features to create a characteristic profile for ERLs, aiding in understanding ERL operational patterns. This will facilitate relevant authorities in formulating targeted control measures and lay the groundwork for further research.
	
	\section{Results: model configuration and tuning}\label{sec:results model}
	
	In this section, we first present the experimental sets and then report the predictive performances of several models. Our goal was to determine the model with the highest predictive performance before investigating the importance of the features.
	
	\begin{figure*}[!t]
	\centering
	\includegraphics[width=0.8\textwidth]{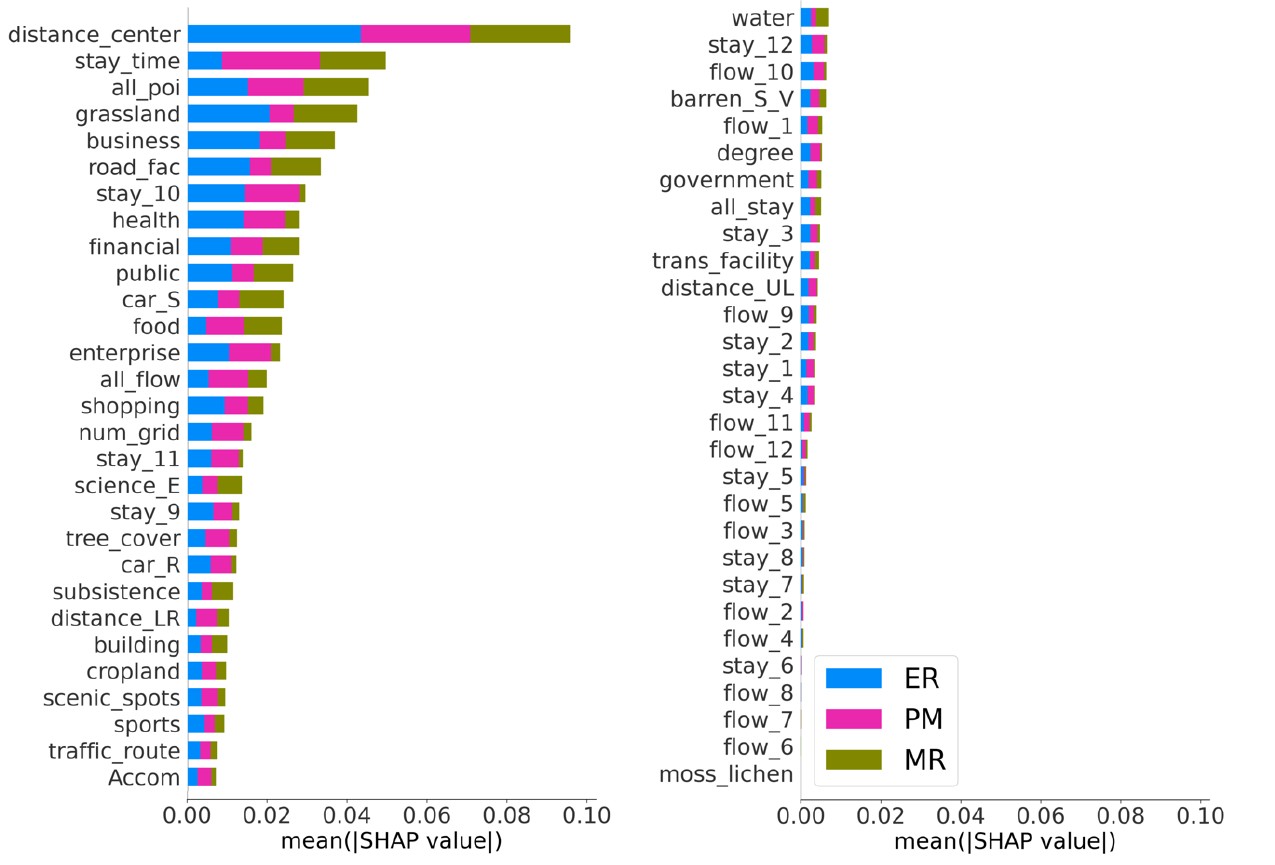}
	\caption{\label{fig:FI_ranking}Feature importance ranking for different categories of samples.}
	\end{figure*}
	
	\begin{figure*}[!t]
		\centering
		\includegraphics[width=\textwidth]{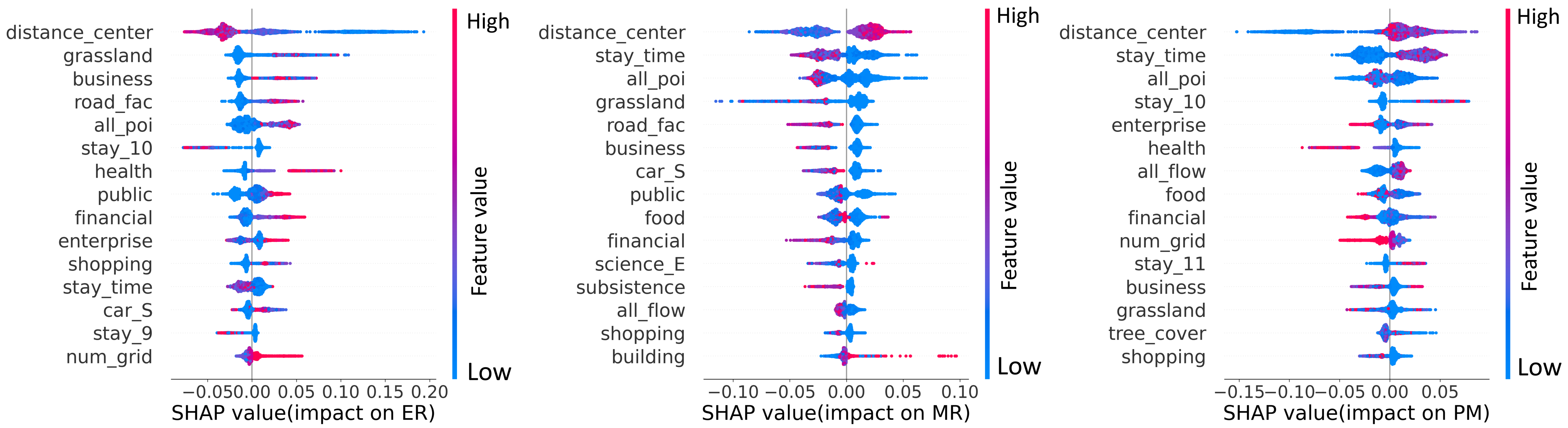}
		\caption{\label{fig:FI_allsample}Feature importance analysis with all samples for three categories.}
	\end{figure*}
	
	\subsection{Experimental setup}
	
	Four classification models were implemented using the Scikit-Learn library in Python. The parameters for LR, MLP, GBDT, and RF were set as follows: for LR, the maximum number of iterations, regularization strength coefficient, and classification weight coefficients were 800, 1.0, and 0.2:0.5:0.3, respectively; for MLP: the activation function, size of the hidden layer, learning rate, and maximum number of iterations were Relu, [256,128], 0.01, 800, and 1.3, respectively; for GBDT: the maximum number of iterations, L2 penalty parameter, learning rate, and maximum depth were 100, 10, 0.01, and 6, respectively; and for RF: the number of trees, maximum depth, and classification weights were 150, 7, and 0.2:0.5:0.3, respectively. To ensure experimental comparability, the same random seeds were applied to all four classification models throughout the experiment. In each experiment, the datasets of the four models were identical.
	
	Detailed information regarding the experimental data is provided in Section~\ref{chap:Performance}. Given that samples from different shifts may belong to the same ERL, the division of the training and validation sets should be based on the 2,403 known ERLs in the ERL registry, rather than on the 20,489 samples over 64 shift times. We randomly selected 70\% as the training set, 10\% as the validation set, and 20\% as the test set from the ERL registry. The model results for the 20\% unseen test set are listed in Table~\ref{tab:Performance}.
	
	\begin{figure*}[!t]
		\centering
		\includegraphics[width=0.8\textwidth]{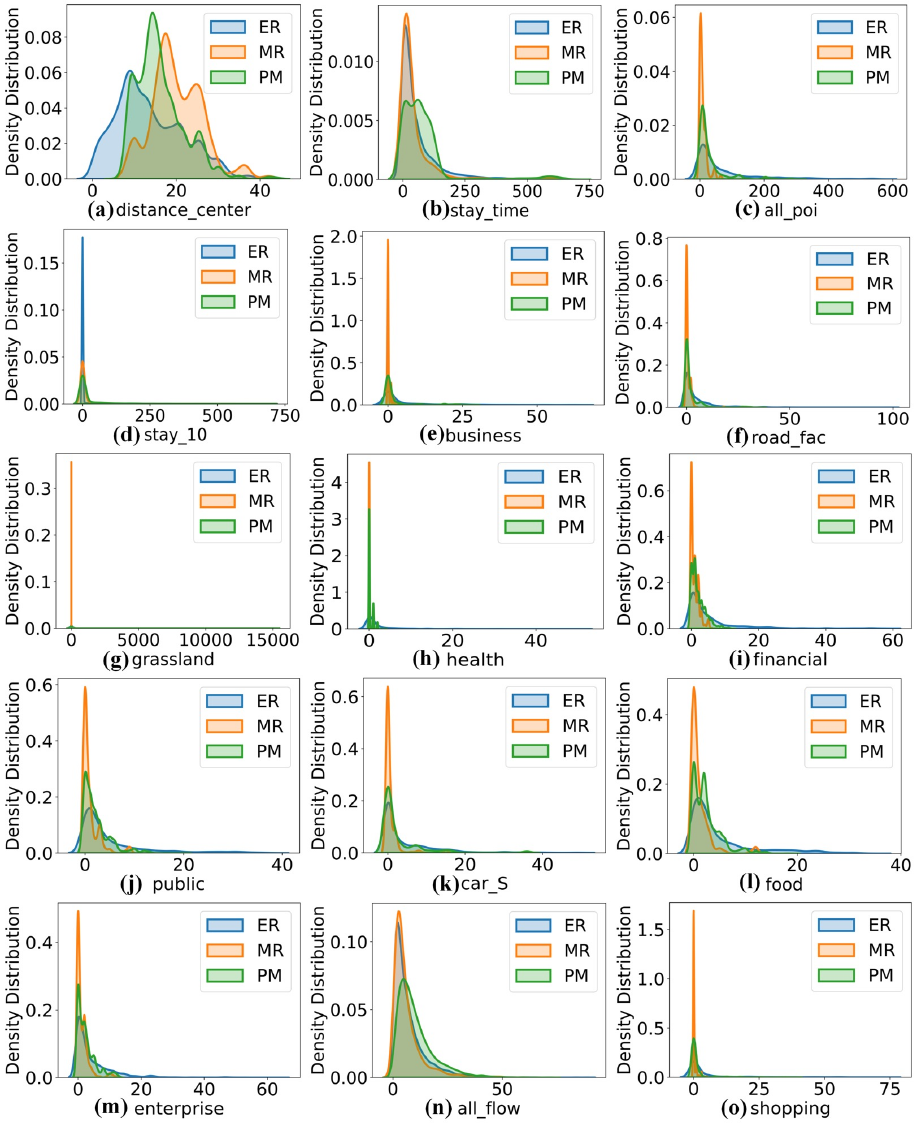}
		\caption{\label{fig:FI_distribution}Distribution of features in different categories.}
	\end{figure*}
	
	\subsection{Performance of different models}\label{chap:Performance}
	
	As far as we know, there is no research foundation for this issue. Therefore, the model's baseline is based on manual judgment. For instance, $\text{the percentage of ERs} = \tfrac{\text{number of ERs}}{\text{number of ERLs}}=65.41\%$ . This means that if all ERLs are predicted to be ERs, only 65.41\% of ERLs are correct even when MR and PM are not considered.
	
	Table~\ref{tab:Performance} presents a comparison of the four models -- LR, RF, MLP, and GBDT -- across the five performance metrics. Of these, LR performs the worst due to the simplicity of the model structure. MLP, GBDT, and RF perform better on all five metrics, well above the baseline (manually estimated). Overall, RF has the best performance, likely due to its superior capability to handle high-dimensional and unbalanced data, and its robustness to data outliers and noise \cite{bernard2009influence}. In addition, The RF model is highly interpretable due to its Gini impurity-based ex-ante interpretation method \cite{bernard2009influence} and SHAP-based ex-post interpretation method, and the SHAP method for tree models has low computational complexity \cite{lundberg2018consistent}. Consequently, we selected RF as the final model for subsequent feature importance experiments.
	
	\begin{figure}[!t]
	\centering
	\includegraphics[width=0.5\textwidth]{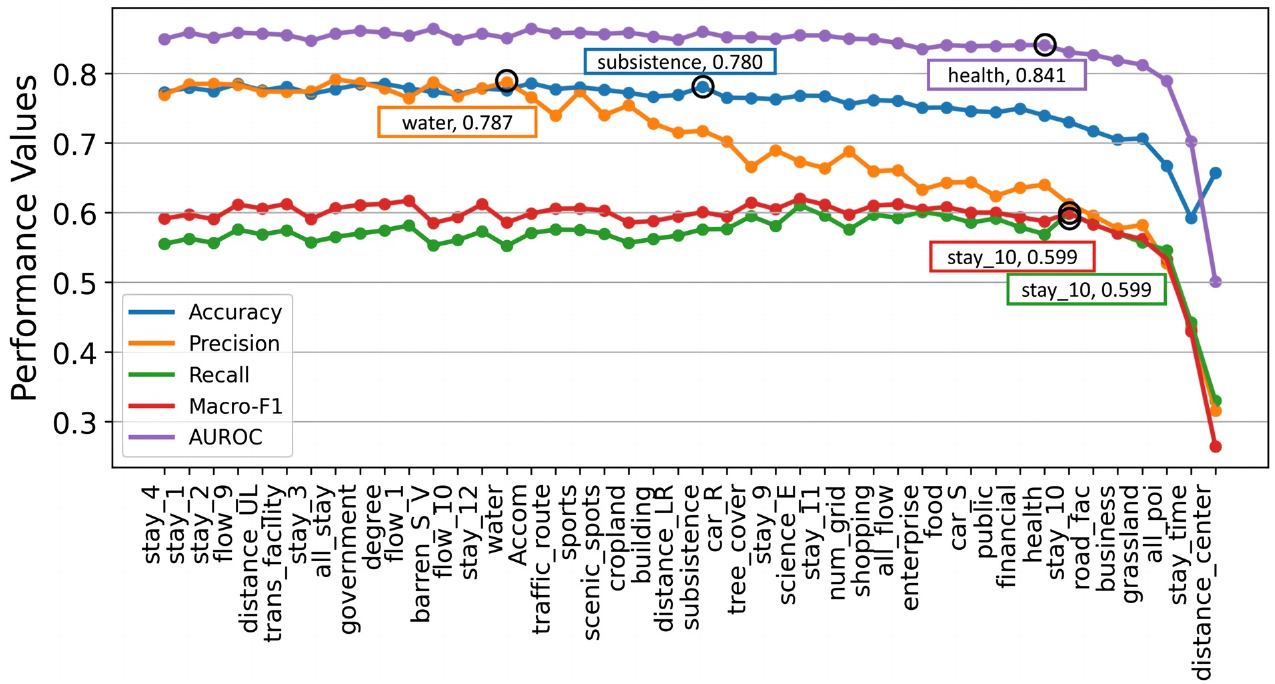}
	\caption{\label{fig:Backward_elimination}Prediction performance for iteratively reducing features until all features are removed (i.e., until random guessing). Black circles denote the prediction performance of the current performance metric with a smaller subset of features.}
	\end{figure}
	
	\begin{figure*}[!h]
	\centering
	\includegraphics[width=0.83\textwidth]{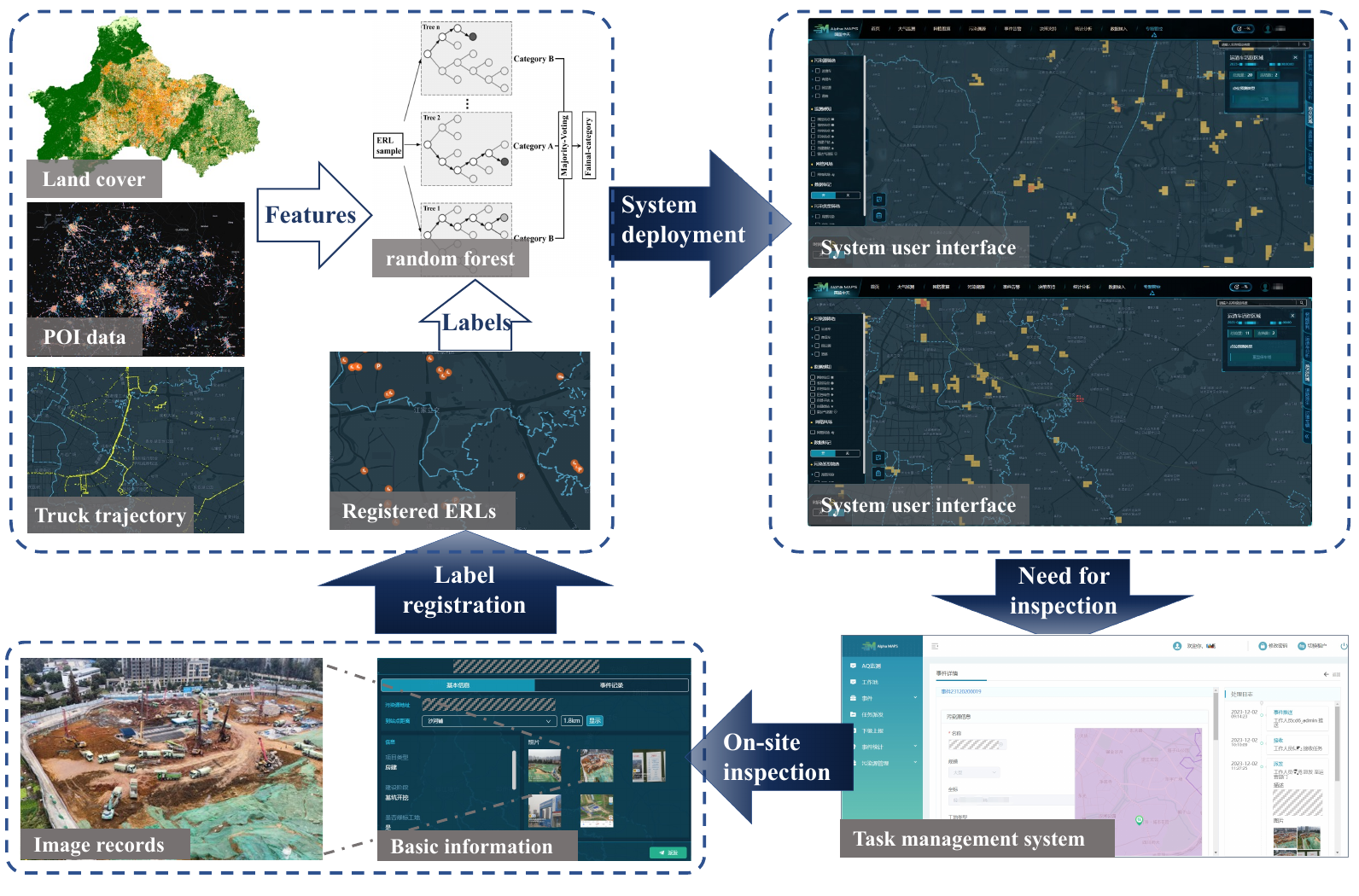}
	\caption{\label{fig:System}Workflow of the ERLs classification algorithm in the Alpha MAPS system.}
	\end{figure*}
	
	\section{Results: feature importance}\label{sec:results feture}
	
	In this section, we examine the relationship between different features and target labels using the SHAP-values and feature distributions. One of our goals is to clarify the importance of features in constructing a more streamlined classification model. The second goal is to mine the operational rules of ERLs so that the relevant authorities can formulate targeted regulatory measures.
	
	\subsection{Feature importance analysis}
	
	As discussed in Section~\ref{sec:Feature importance}, extracting the feature importance from classification models is a crucial process. Fig.~\ref{fig:FI_ranking} depicts the importance ranking of 58 features in the RF model as determined by the SHAP-values. In Fig.~\ref{fig:FI_ranking}, the vertical axis represents various features and the horizontal axis represents the sum of the absolute SHAP values for all samples—the feature importance values. The sum of the importance values for all the features was normalized to 1. For a more comprehensive understanding and evaluation of the importance values, Fig.~\ref{fig:FI_allsample} summarizes the SHAP-values for the top 15 features, each with an importance value exceeding 0.02. Each data point represents a sample, where blue indicates smaller feature values and red indicates larger feature values. The relationship between the sample and SHAP values is depicted in Fig.~\ref{fig:FI_allsample}. To provide a more intuitive understanding of the differences in the feature values among the different categories, Fig.~\ref{fig:FI_distribution} shows the numerical distributions of the top 15 features.
	
	The $distance\_center$ in the basic geographic features, representing the distance of the samples from the city center, is the most crucial for predictions across different categories. This is particularly evident when predicting ER, where the importance value of $distance\_center$ is notably higher than that of the other features. This observation is consistent with the results presented in Figs.~\ref{fig:FI_allsample} and~\ref{fig:FI_distribution} (a), that is, as the sample proximity to the city center increases, the probability of it belonging to ER is higher, followed by MR, and finally PM. This pattern is likely attributable to the distribution of construction sites in ER, which are typically spread across the city. By contrast, MR sites (quarry, commercial concrete mixing station) and parking lots are constrained from being located in the city center. Conversely, parking lots tend to be situated farther away from the city center. This conclusion aligns with urban planning principles \cite{peng2015studies}.
	
	Among the transport related features, $stay\_time$, $all\_flow$ (total inflow within one shift), and $stay\_10$ (total number of stay points in the 10th hour of a shift) demonstrated higher importance values. As shown in Figs.~\ref{fig:FI_allsample} and~\ref{fig:FI_distribution} (b,g,n), for $stay\_time$, the main reason is that CWHTs stay in the parking lot for a longer period. For $all\_flow$, this may be attributed to the fact that the flow of CWHTs in construction sites, quarries, and commercial mixing concrete stations is typically determined by the workload; therefore, the inflow is generally not very high. However, parking lots often accommodate CWHTs from other ERLs, resulting in relatively large flows. For $stay\_10$, the 10th hour of a shift typically marks the end of the working day, and CWHTs primarily transition from work areas to parking lots. Consequently, the $stay\_10$ in parking lots is larger, $stay\_10$ for ER is smaller, and for MR-possibly due to the smaller amount of data—the feature importance of $stay\_10$ is lower. Interestingly, the importance value of $stay\_10$ exceeds that of $flow\_10$. As discussed in Section~\ref{sec:Transport related features}, although the physical meaning of $stay\_10$ is not as explicit as that of $flow\_10$, it may contain additional information. Fig.~\ref{fig:FI_ranking} also indicates that the feature importance values of $flow\_t\;(t=1,2,\dots,12)$ are generally low. This may be attributed to the inherent randomness in the operations of CWHTs, which lack unified and effective regulation. This may be attributed to the strong randomness in the operations of CWHTs, which lacks unified and effective regulation during the working period. The typical operating modes are ``first come, first load", ``first come, first unload", and ``arrive later, queue up", and evaluating the travel and queue times on the road is challenging.
	
	Within the land cover feature, $grassland$ exhibited a high importance value, whereas the remaining features demonstrated a lower performance.  As shown in Figs.~\ref{fig:FI_allsample} and~\ref{fig:FI_distribution} (d), samples with a larger $grassland$ feature were more likely to be predicted as MR and, conversely, as in other categories. This may be because the earth dumping ground in ER is usually located in the outer suburbs and the land surface is not treated, thus preserving the original $grassland$. By contrast, quarries, commercial concrete mixing stations, and parking lots are also located in the outer suburbs, but their land surfaces are usually treated, and almost no $grassland$ is preserved. Fig.~\ref{fig:FI_ranking} indicates that several other land cover features have lower importance values. Previous studies \cite{chen2020classification, lynch2020classification} have successfully used urban satellite remote sensing data to identify and classify urban locations. The relatively low importance values of land-cover features in this study suggest that structured land-cover data may not capture sufficient information to compete with remotely sensed image data.
	
	The importance values of the POI features are generally high, particularly for features such as $all\_poi$, $business$, $road\_fac$, $health$. As observed in Figs.~\ref{fig:FI_allsample} and~\ref{fig:FI_distribution}, the number of POI in each category around MR is the smallest, followed by PM, and finally ER. Although we categorized POI into 17 categories, numerous other criteria for categorizing POI exist, and other categories of POI data are worth considering.
	
	\subsection{Model simplification}
	Dependencies exist among the features, and the removal of certain features leads to changes in the importance values of the remaining features. Specifically, as illustrated by the ordering of importance values in Fig. ~\ref{fig:FI_ranking}, features such as $moss\_lichen, flow\_6,\dots, flow\_11$ exhibit low importance values; therefore, these features were simultaneously removed. Then, $stay\_4, stay\_1,\dots, stay\_time, distance\_center$ was removed and the changes in the values of the five performance metrics were dynamically recorded; the results are shown in Fig.~\ref{fig:Backward_elimination}.
	
	In Fig.~\ref{fig:Backward_elimination}, the horizontal axis, which progresses from left to right, represents the gradual removal of features, and the vertical axis denotes the values of the model performance metrics. The points marked by black circles indicate instances where the model performance begins to decline significantly, with slight variations in the points at which different performance metrics decrease. The precision initially experiences a significant drop, whereas the recall exhibits a corresponding slow upward trend. In addition, Macro-F1 remains almost unchanged, making it suitable for practical applications. Many studies regard Macro-F1 as a measure of the comprehensive performance of the model; therefore, our RF model only needs to retain the six features $distance\_center$, $stay\_time$, $all\_poi$, $grassland$, $business$, and $road\_fac$ to achieve a high level of performance.
	
	\section{Real-world deployment}\label{sec:system}
	The ERLs classification algorithm was deployed in November 2023 in the {\it Alpha MAPS} system, for smart air pollution monitoring and management in Chengdu. The {\it Alpha MAPS} was implemented in Chengdu (a megacity in China) starting in March 2022, and has played a significant role in urban environmental monitoring and governance.

	The proposed algorithm's functioning within the {\it Alpha MAPS} system is illustrated in Fig. \ref{fig:System}. For each operation, the ERL classification algorithm utilizes dump truck trajectories spanning the previous 61 days (61 day shifts and 61 night shifts), POI data, land cover data, and the ERL registry, to classify ERLs on that shift. These locations are subsequently ranked according to risk, considering their category, network relationships, CWHTs flow, and stay time. High-risk ERLs are promptly posted to inspectors in real-time. Inspectors will visit ERL sites to record pertinent information, including images, categories, sizes, operational status, and whether environmental activities are underway et al. This information serves a dual purpose: it is uploaded to the {\it Alpha MAPS} system to update ERLs information and increase label number, while also being transmitted to government departments to facilitate law enforcement decision-making.
	
	During December 2023, the module has performed classification tasks for 16,132 ERLs with a classification accuracy of 77.8\% based on 1,092 on-site verifications. These include 724 construction cites/earth dumping ground, 48 concrete mixing stations, and 80 truck parking locations. The ERLs classification functionality has considerably enhanced local authorities' capacity to monitor and manage locations with high dust pollution risks. 
	
	\section{Conclusion}\label{sec:conclusion}
	This study aimed to develop a classification model for urban dust pollution sources by introducing a set of features that could potentially influence dust pollution sources. These features include basic geographic, land cover, urban POI, and transport related features. Using real data from Chengdu City, we aimed to establish the most suitable classification model for urban dust pollution sources. Comparative experiments involving four machine learning models revealed that the RF model outperformed the other models regarding classification effectiveness.
	
	A key aspect of this study is the identification of feature importance and model simplification. The feature importance identification relies on the SHAP-value method, whereas model simplification combines the SHAP-value and backward elimination methods. Quantitative analysis indicated that $distance\_center$ (the distance of samples from the city center) was the most crucial feature, followed by $stay\_time$, $all\_poi$ (number of all POI within a 1km radius), and $grassland$ (area ratio of low herbaceous). Furthermore, feature importance analysis not only aids in comprehending the operational patterns of dust pollution sources but also guides model simplification. The results of the model simplification demonstrate that high-accuracy classification can be achieved using only a small subset of features, including $distance\_center$, $stay\_time$, $all\_poi$, $grassland$, $business$, and $road\_fac$.
	
	Future research on urban dust pollution sources can progress in several directions. (1) More urban data can be gathered and the classification model can be tailored to meet the environmental regulatory needs of different cities within the framework of this study. (2) Additional machine learning methods can be considered, such as lasso regression and support vector machines, and deep learning methods can be explored. (3) The feature set can be expanded to encompass the multifaceted characteristics of dust pollution sources. (4) A model for evaluating and controlling urban dust pollution sources can be constructed by combining active time and in/outflow data for a comprehensive assessment and ranking of dust risks. We acknowledge that despite the considerable progress made in this study, there is still potential for further improvement in the model accuracy, and pursuing these directions may prove to be an effective means of achieving this goal.

	\vspace{11pt}

	\begin{IEEEbiography}[{\includegraphics[width=1in,height=1.25in,clip,keepaspectratio]{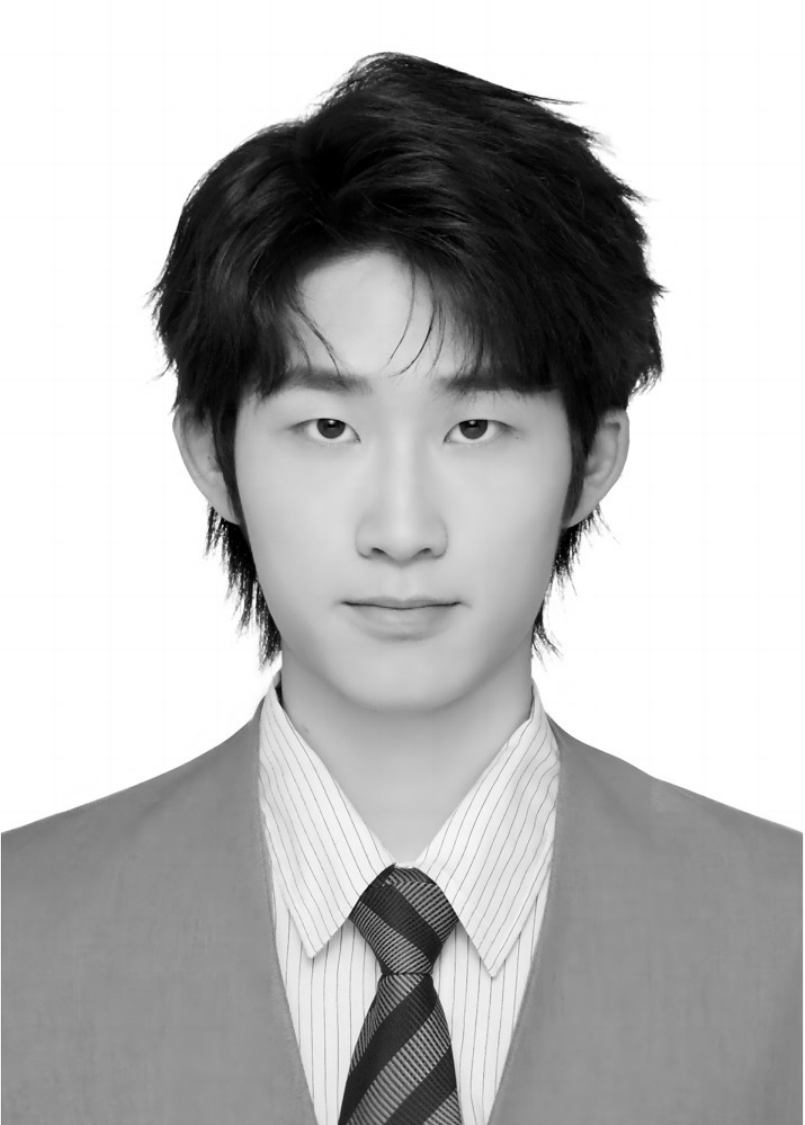}}]{Lei Yu}
		received the B.S. degree in transportation from Southwest Jiaotong University, Chengdu, China, in 2023. He is currently pursuing the M.S. degree with the School of Transportation and Logistics at Southwest Jiaotong University, Chengdu, China. His research interests include urban computing, traffic modeling, and operations research.
	\end{IEEEbiography}
	
	\begin{IEEEbiography}[{\includegraphics[width=1in,height=1.25in,clip,keepaspectratio]{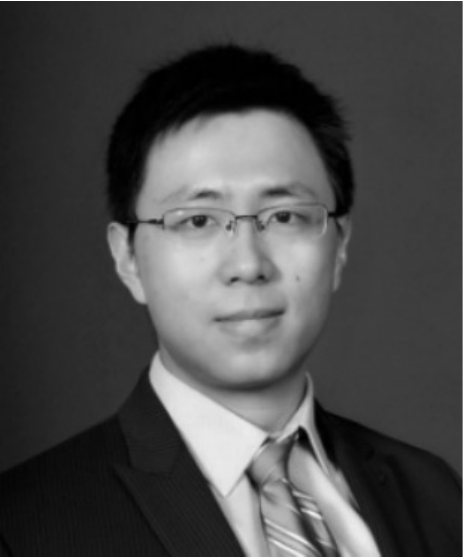}}]{Ke Han}
	received the B.S. degree in mathematics from the University of Science and Technology of China in 2008, and the Ph.D. degree in mathematics from Pennsylvania State University in 2013. He was appointed as a Lecturer in 2013, and then a Senior Lecturer in 2018, with the Centre for Transport Studies, Department of Civil and Environmental Engineering, Imperial College London. He joined Southwest Jiaotong University, China, as a Full Professor in 2020. He has published two book and over 140 journal and conference papers in the areas of transportation modeling and optimization, and sustainable urban management.

	\end{IEEEbiography}
	
		\vfill

\end{document}